\documentclass[conference]{IEEEtran}
\IEEEoverridecommandlockouts
\usepackage{cite}
\usepackage{amsmath,amssymb,amsfonts}
\usepackage{algorithmic}
\usepackage{graphicx}
\usepackage{textcomp}
\usepackage{xcolor}
\usepackage{booktabs} 

\newcommand{\affmark}[1]{\raisebox{0.6ex}{\scriptsize\textbf{#1}}}

\def\BibTeX{{\rm B\kern-.05em{\sc i\kern-.025em b}\kern-.08em
    T\kern-.1667em\lower.7ex\hbox{E}\kern-.125emX}}
\begin{document}

\title{Season: Spectrum-Aware Orthogonal Gradient Refinement for Transfer-Based Adversarial Attacks}

\author{
\makebox[\textwidth][c]{%
\parbox[t]{0.25\textwidth}{\centering
\textbf{Tianyi Wang\affmark{1}}\\
\textit{\affmark{1} Tongji University}\\
Shanghai, China\\
\scriptsize 2351270@tongji.edu.cn
}\hfill
\parbox[t]{0.31\textwidth}{\centering
\textbf{Zhenghao Gao\affmark{1,2}}\\
\textit{\affmark{1} Huazhong University of Science and Technology}\\
\scriptsize \textit{\affmark{2} Wuhan LightRead Intelligent Technology Co., Ltd.}\\
Wuhan, China\\
\scriptsize u202215226@hust.edu.cn\\
\scriptsize gaozhenghao@lightingread.cn
}\hfill
\parbox[t]{0.31\textwidth}{\centering
\textbf{Shengjie Xu\affmark{1,2,*}}\\
\textit{\affmark{1} Huazhong University of Science and Technology}\\
\scriptsize \textit{\affmark{2} Wuhan LightRead Intelligent Technology Co., Ltd.}\\
Wuhan, China\\
\scriptsize u202317280@hust.edu.cn\\
\scriptsize xushengjie@lightingread.cn
}%
}
\thanks{\textsuperscript{*}Corresponding author: Shengjie Xu.}
}

\maketitle
\vspace{-0.9\baselineskip}

\begin{abstract}
Transfer-based adversarial attacks often transfer poorly across heterogeneous architectures because CNNs favor local textures while Vision Transformers (ViTs) rely on global shapes. We propose Season, a spectrum-aware orthogonal gradient refinement framework for L-infinity transfer attacks against black-box target models on ImageNet, using a white-box surrogate. Season decomposes each update into a low-frequency branch capturing structural cues and a high-frequency branch capturing textures. A low-saliency guidance scheme reallocates high-frequency energy to background regions, preserving foreground structures that ViTs depend on. An orthogonal projection then forces the textural update to lie in the orthogonal complement of the structural direction, mitigating feature interference. As a training-free plug-and-play wrapper, Season enhances eight gradient-stabilization and input-enhancement attacks without modifying their cores. Across eight CNN, ViT, and MLP targets, Season improves transfer success rate by 6.6 percentage points on average and up to 16.0 points over strong baselines under a unified protocol.
\end{abstract}

\begin{IEEEkeywords}
Adversarial Attacks, Transferability, Spectrum Analysis, Vision Transformers, Orthogonal Refinement
\end{IEEEkeywords}

\section{Introduction}
\label{sec:intro}

\begin{figure*}[t]
\centering
\includegraphics[width=0.94\textwidth]{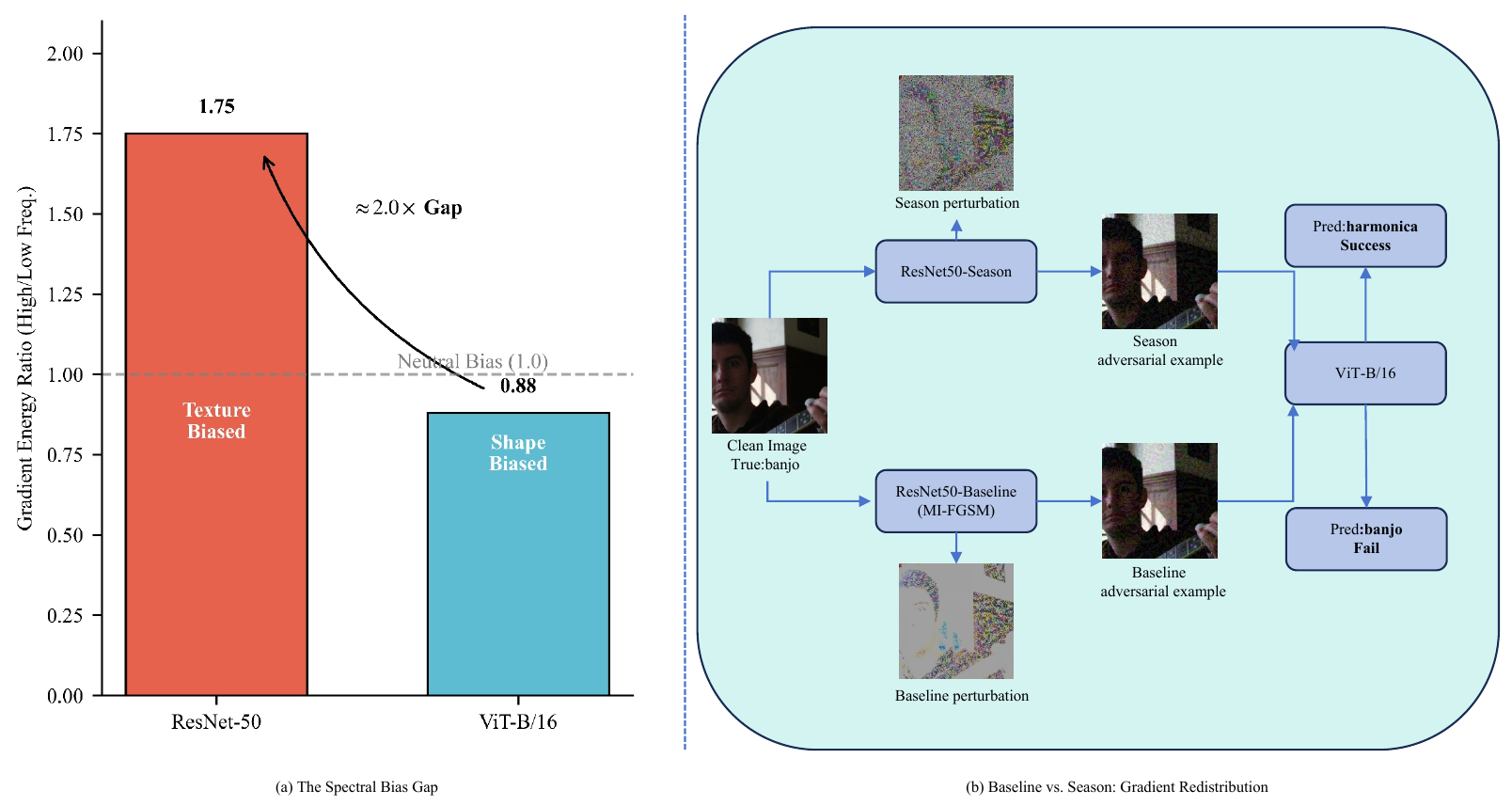}
\caption{(a) Gradient spectrum analysis reveals a $2\times$ spectral bias gap between CNN and ViT: ResNet-50 exhibits a high/low frequency energy ratio of 1.75, while ViT-B/16 shows only 0.88. (b) Season redirects high-frequency perturbations to low-saliency background regions while preserving foreground structures essential for ViT transferability.}
\label{fig:teaser}
\end{figure*}

Deep neural networks are well known to be vulnerable to adversarial examples, which can cause confident misclassification under imperceptible perturbations~\cite{goodfellow2014explaining}. Among different threat models, \emph{transfer-based} attacks are particularly practical because they only require access to a local surrogate model and rely on the empirical transferability of adversarial examples to unseen targets~\cite{papernot2017practical}. A long line of work has therefore focused on improving transferability by stabilizing gradients (e.g., momentum) or enriching input transformations (e.g., diversity)~\cite{dong2018boosting,xie2019improving}. Recent systematization studies categorize these methods and report strong performance when attacking models within the same architectural family, especially CNN-to-CNN transfer~\cite{zhao2025revisiting}. However, their effectiveness degrades substantially when attacking heterogeneous targets such as Vision Transformers (ViTs) and MLP-like architectures, revealing a persistent \emph{cross-architecture transfer gap}.
We argue that a key reason for this failure lies in the \textbf{spectral bias mismatch} between architectures. ImageNet-trained CNNs are known to exhibit a strong bias towards local textures~\cite{geirhos2018imagenet}, while ViTs, enabled by global self-attention, tend to rely more on holistic shapes and long-range structures~\cite{dosovitskiy2020image,naseer2021intriguing}. Our empirical gradient spectrum analysis on ImageNet (Fig.~\ref{fig:teaser}(a)) confirms this discrepancy quantitatively: the ratio between high- and low-frequency gradient energy is \textbf{1.75} for a ResNet-50 surrogate but only \textbf{0.88} for a ViT-B/16, indicating almost a \textbf{2$\times$ gap} in spectral preference. Standard transfer-based attacks, which greedily follow the raw surrogate gradient, therefore inject excessive high-frequency noise aligned with the CNN's texture bias. This not only risks overfitting to the source model, but also corrupts the low-frequency structural cues that are crucial for fooling ViT-like targets. Recent benchmarks further show that many transfer attacks implicitly favor either CNNs or ViTs, but rarely both~\cite{zhao2025revisiting}.

To bridge this gap, we propose \textbf{Season} (Spectrum-Aware Orthogonal Gradient Refinement), a plug-and-play framework that explicitly coordinates texture and shape cues during optimization. Season first performs a \textbf{dual-frequency decomposition} of the surrogate gradient into a low-frequency branch capturing structural information and a high-frequency branch capturing textural details. Then, motivated by the fact that ViTs rely heavily on foreground structures while CNNs are sensitive to local textures, we introduce a \textbf{low-saliency guidance} strategy that redirects high-frequency perturbations towards low-saliency background regions while attenuating updates on salient objects (Fig.~\ref{fig:teaser}(b)). Finally, Season enforces a \textbf{geometric orthogonal projection} between the two branches, preventing aggressive textural updates from interfering with the structural direction. The resulting update simultaneously enhances CNN-oriented texture corruption and preserves ViT-oriented shape information.

Our main contributions are summarized as follows:
\begin{itemize}
    \item \textbf{Spectrum-aware spatial-spectral decoupling.} Instead of treating the gradient as a monolith, we propose a novel decomposition strategy that explicitly separates dual-frequency branches. We further introduce \textbf{low-saliency guidance} to redirect high-frequency perturbations towards background regions, exploiting CNN texture bias without disrupting the foreground structures essential for ViTs.
    \item \textbf{Geometric orthogonal refinement.} We introduce a geometric \textbf{orthogonal projection} constraint to resolve the feature interference between decoupled branches. This mechanism forces textural updates to evolve in the null space of the structural direction, ensuring that high-frequency noise does not corrupt low-frequency semantic cues.
    \item \textbf{Strong cross-architecture transferability.} As a generic plug-and-play module, Season consistently boosts eight representative transfer attacks. Extensive experiments on ImageNet show an average improvement of \textbf{6.6\%} (up to \textbf{16.0\%}) over strong baselines against heterogeneous CNN, ViT, and MLP targets.
\end{itemize}

\section{Related Work}
\label{sec:related}

\subsection{Gradient Stabilization and Input Enhancement}
Following the taxonomy in~\cite{zhao2025revisiting}, most practical transfer-based attacks fall into two families: gradient stabilization and input enhancement. Gradient-stabilization methods aim to escape local optima of the surrogate model by smoothing or pre-conditioning the update direction. Beyond the momentum-based MI-FGSM~\cite{dong2018boosting}, \textbf{NI-FGSM}~\cite{lin2019nesterov} incorporates Nesterov acceleration to look ahead along the momentum direction, and \textbf{PI-FGSM}~\cite{gao2020patch} amplifies gradients on patch-wise regions to reinforce local cues. \textbf{VMI-FGSM}~\cite{wang2021enhancing} further explicitly reduces gradient variance by aggregating gradients over a local neighborhood.

Input-enhancement methods improve generalization by diversifying input patterns. \textbf{DIM}~\cite{xie2019improving} applies random resizing and padding. \textbf{TIM}~\cite{dong2019evading} convolves gradients with a pre-defined kernel to approximate translation invariance. \textbf{SIM}~\cite{lin2019nesterov} averages gradients over multi-scale copies of the input. More recently, \textbf{VT}~\cite{wang2021enhancing} samples in the vicinity of the input to reduce gradient variance, and \textbf{Admix}~\cite{wang2021admix} mixes the input with images from other categories to encourage category-level generalization. While these methods are effective, they still optimize within a \emph{single, coupled gradient space} dominated by the surrogate's bias (e.g., texture for CNNs), which limits their transferability to shape-biased architectures such as ViTs. Season is designed as a universal wrapper that refines the gradients produced by these attacks and is conceptually complementary rather than competitive.

\subsection{Frequency-aware and Feature-based Attacks}
To move beyond purely pixel-space constraints, several works manipulate gradients in the frequency or feature domain. \textbf{SGM}~\cite{wu2020skip} and \textbf{LinBP}~\cite{guo2020backpropagating} argue that low-frequency components are more transferable and modify the backward pass (e.g., skipping residual branches or linearizing activations) to attenuate high-frequency noise. However, these methods rely on heuristic network-specific modifications and tend to implicitly suppress a large portion of high-frequency content.

Feature-based attacks, such as \textbf{FIA}~\cite{wang2021feature} and \textbf{NAA}~\cite{zhang2022improving}, instead disrupt intermediate feature maps to break semantic representations. Although effective, they require choosing target layers and aggregating features across them, which incurs additional implementation complexity and runtime overhead. In contrast, Season operates directly on the input gradient with only a small constant overhead on top of the base attack, and—unlike SGM-style approaches—explicitly \emph{decouples} rather than discards high-frequency signals, leveraging them to attack background textures via our saliency-guided mask. Beyond transfer attacks, gradient decoupling has also been explored to accelerate saliency-map based attacks; FastJSMA decomposes Jacobian computation into complementary gradient components to significantly reduce the computational cost~\cite{gao2025fastjsma}.

\begin{figure*}[t]
\centering
\includegraphics[width=0.94\textwidth]{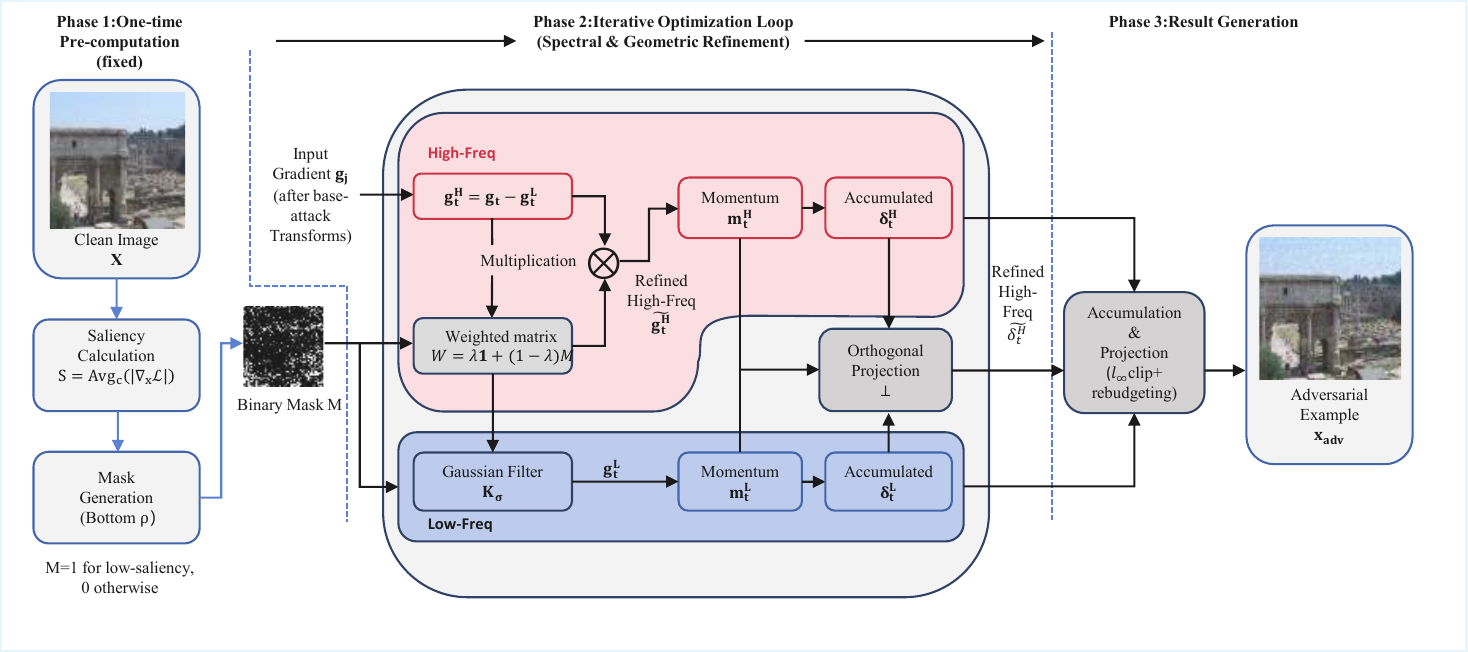}
\caption{Overview of the Season framework. (Left) Dual-frequency decomposition separates the gradient into low-frequency structural and high-frequency textural branches. (Middle) Low-saliency guidance redirects textural perturbations to background regions. (Right) Orthogonal projection ensures the two branches evolve independently without interference.}
\label{fig:framework}
\end{figure*}

\section{Methodology}
\label{sec:method}

\subsection{Problem Formulation and Gradient Coupling}
\label{sec:problem}

We consider the standard untargeted transfer attack setting under the $\ell_\infty$norm constraint, where the target model(s) are treated as black-box (no gradient or query access).
Given a clean image $x$ with ground-truth label $y$ and a white-box surrogate model $f_s$, the goal is to find a perturbation $\delta \in \mathcal{B}_\epsilon(x)$ that maximizes the loss 
\begin{equation}
    \max_{\delta \in \mathcal{B}_{\epsilon}(x)} \; \mathcal{L}\big(f_s(x+\delta), y\big),
\end{equation}
where $\mathcal{B}_{\epsilon}(x)$ denotes the $\ell_{\infty}$ ball of radius $\epsilon$ around $x$.  
Iterative transfer attacks such as MI-FGSM and its transformed variants (e.g., DIM, TIM) update a momentum buffer $m_t$ using the gradient $g_t$ and then perform
\begin{equation}
    x_{t+1} = \Pi_{\mathcal{B}_{\epsilon}}\!\big(x_t + \alpha \cdot \mathrm{sign}(m_t)\big),
\end{equation}
where $\alpha$ is the step size and $\Pi_{\mathcal{B}_{\epsilon}}$ denotes projection onto the feasible set.

In these standard formulations, the update direction $m_t$ aggregates all components of $g_t$ into a single stream, implicitly coupling texture- and shape-related signals.  
As suggested by prior analyses on ImageNet-trained CNNs~\cite{geirhos2018imagenet} and our spectrum study in Fig.~\ref{fig:teaser}(a), CNN loss gradients are empirically dominated by high-frequency textures.  
When such coupled gradients are directly followed, texture-biased components tend to overshadow low-frequency structural signals, leading to poor transferability on shape-biased architectures such as Vision Transformers (ViTs).  
To address this issue, we propose \textbf{Season}, a framework that explicitly decouples and coordinates these conflicting components.

\subsection{Season: Spectrum-Aware Decomposition and Guidance}
\label{sec:season_core}

Season refines the raw surrogate gradient $g_t$ through a coordinated spectral–spatial process.

\paragraph{Dual-Frequency Decomposition}
To separate structural and textural information, we adopt a Gaussian-based decomposition that preserves \emph{spatial locality} compared to hard frequency truncation.  
Let $K_{\sigma}$ be a Gaussian kernel with scale $\sigma$.  
Given the base gradient $g_t$, we compute a low-frequency component and its high-frequency residual:
\begin{equation}
    g_t^{L} = K_{\sigma} * g_t, \quad g_t^{H} = g_t - g_t^{L},
\end{equation}
where $*$ denotes convolution.  
In practice, each branch is further normalized (e.g., by its mean absolute value) before being fed into the momentum buffers, so that subsequent updates depend mainly on their \emph{spatial patterns} rather than absolute scale.  
Intuitively, $g_t^{L}$ captures global, low-frequency structures that are more robust across architectures, while $g_t^{H}$ concentrates local, high-frequency textures that are more specific to CNN surrogates.

\paragraph{Precomputed Low-Saliency Guidance}
Blindly amplifying $g_t^{H}$ can disrupt semantic structures on the foreground object.  
We therefore steer texture perturbations toward non-robust background regions.  
To limit additional overhead and avoid misalignment with input transformations, we compute a saliency map
\begin{equation}
S_{ij} = \frac{1}{C}\sum_{c=1}^{C}\left|\frac{\partial \mathcal{L}\!\left(f_s(x),\,y\right)}{\partial x_{cij}}\right| ,
\end{equation}

on the \emph{clean image} $x$ using a single backward pass, and keep it fixed throughout optimization.  
In implementation, we use the channel-wise gradient magnitude (mean absolute value across channels) as the saliency score.  

We then select the bottom-$\rho$ percentile (default $\rho = 0.15$) of pixels in $S$ to form a binary \emph{low-saliency mask} $M$, where $M_{ij}=1$ indicates low-saliency (background) locations.  
A static spatial weight matrix is constructed as
\begin{equation}
    W = \lambda \mathbf{1} + (1-\lambda) M,
\end{equation}
where $\mathbf{1}$ is an all-ones matrix and $\lambda \in [0,1]$ controls the attenuation on high-saliency regions.  
The high-frequency branch is re-weighted as
\begin{equation}
    \tilde{g}_t^{H} = g_t^{H} \odot W,
\end{equation}
with $\odot$ denoting element-wise multiplication.  
This static weighting biases the subsequent high-frequency momentum to accumulate predominantly in low-saliency background regions, while reducing its contribution on salient foreground objects.  
As a result, Season better exploits CNN texture bias and alleviates unnecessary disturbance to the structural cues that are critical for ViT transferability.

\begin{table*}[t]
\centering
\caption{Transfer attack success rate (\%) of baseline attacks and their Season-enhanced variants. Source model: ResNet-50. Best results per column in \textbf{bold}.}
\label{tab:main_results}
\resizebox{\textwidth}{!}{%
\begin{tabular}{l|c|ccccc|ccc|c}
\hline
\textbf{Method} & \textbf{Source} & \textbf{RN-101} & \textbf{DN-121} & \textbf{VGG-19} & \textbf{MB-V2} & \textbf{GoogLeNet} & \textbf{ViT-B/16} & \textbf{Swin-T} & \textbf{Mixer-B} & \textbf{Avg.} \\
\hline
MI & 99.97 & 90.4 & 85.24 & 77.39 & 76.69 & 67.67 & 29.26 & 45.73 & 37.2 & 63.7 \\
Season + MI & 99.97 & 93.21 & 87.95 & 81.81 & 81.61 & 71.54 & 34.76 & 49.43 & 41.9 & 67.78 \\
\hline
PI & 99.97 & 92.27 & 87.08 & 79.99 & 79.37 & 69.97 & 29.95 & 46.8 & 38.24 & 65.46 \\
Season + PI & 99.97 & 94.23 & 88.91 & 83.39 & 83.22 & 72.74 & 35.08 & 50.13 & 42.18 & 68.73 \\
\hline
TI & 99.92 & 75.22 & 72.29 & 64.66 & 67.44 & 53.33 & 29.29 & 31.61 & 34.44 & 53.53 \\
Season + TI & 99.96 & 93.02 & 88.59 & 81.13 & 81.84 & 72.99 & 40.37 & 49.12 & 45.01 & 69.01 \\
\hline
SI & 97.57 & 83.45 & 76.48 & 70.00 & 71.65 & 62.03 & 30.57 & 38.59 & 36.37 & 58.64 \\
Season + SI & \textbf{100.0} & 95.76 & 95.23 & 88.17 & 88.73 & 87.68 & 40.45 & 50.69 & 50.01 & 74.59 \\
\hline
VT & 98.31 & 93.23 & 90.31 & 85.94 & 85.31 & 80.86 & 48.41 & 60.03 & 51.68 & 74.47 \\
Season + VT & 99.98 & 98.67 & 97.2 & 95.23 & 94.65 & 91.97 & \textbf{55.84} & 67.27 & \textbf{61.81} & 82.83 \\
\hline
Admix & 99.93 & 95.69 & 92.85 & 86.99 & 85.57 & 80.67 & 38.74 & 58.56 & 46.53 & 73.2 \\
Season + Admix & 99.99 & 98.15 & 95.4 & 92.24 & 91.52 & 84.92 & 42.17 & 59.91 & 50.46 & 76.85 \\
\hline
NI & 99.97 & 94.55 & 89.25 & 83.64 & 82.68 & 73.47 & 30.5 & 48.49 & 38.95 & 67.69 \\
Season + NI & 99.97 & 94.96 & 88.54 & 85.21 & 84.8 & 74.06 & 33.73 & 49.03 & 42.08 & 69.05 \\
\hline
DI & 99.97 & 98.44 & 98.21 & 95.97 & 94.84 & 94.14 & 53.35 & 69.85 & 58.24 & 82.88 \\
Season + DI & 99.98 & \textbf{98.9} & \textbf{98.56} & \textbf{96.57} & \textbf{96.02} & \textbf{94.83} & 55.56 & \textbf{70.45} & 59.77 & \textbf{83.83} \\
\hline
\end{tabular}%
}
\end{table*}

\subsection{Orthogonal Refinement and Plug-and-Play Integration}
\label{sec:season_integration}

\paragraph{Orthogonal Projection in Perturbation Space}
Simply summing $g_t^{L}$ and $\tilde{g}_t^{H}$ (or their momenta) may re-introduce interference between texture and structure.
Season therefore maintains distinct momentum states $(m_t^{L}, m_t^{H})$ and corresponding perturbation components $(\delta_t^{L}, \delta_t^{H})$.
At each iteration, we first update the dual momentum buffers with decay $\mu$:
\begin{equation}
    m_t^{L} = \mu m_{t-1}^{L} + g_t^{L}, \quad
    m_t^{H} = \mu m_{t-1}^{H} + \tilde{g}_t^{H},
\end{equation}
and obtain their signed update directions.
The low- and high-frequency perturbations are then incremented as
\begin{equation}
    \delta_t^{L} = \delta_{t-1}^{L} + \alpha \cdot \mathrm{sign}(m_t^{L}), \quad
    \delta_t^{H} = \delta_{t-1}^{H} + \alpha \cdot \mathrm{sign}(m_t^{H}).
\end{equation}

Since structural cues act as the shared ``anchor'' across architectures, we project the accumulated high-frequency perturbation onto the orthogonal complement of the structural component in the perturbation space:
\begin{equation}
    \hat{\delta}_t^{H} = \delta_t^{H} -
    \frac{\langle \delta_t^{H}, \delta_t^{L} \rangle}{\|\delta_t^{L}\|_2^2 + \xi}\,\delta_t^{L},
\end{equation}
where $\xi$ is a small constant for numerical stability.
This yields $\langle \hat{\delta}_t^{H}, \delta_t^{L} \rangle \approx 0$, which empirically reduces coupling between textural and structural updates.
The combined perturbation is $\delta_t = \delta_t^{L} + \hat{\delta}_t^{H}$.
To enforce the $\ell_{\infty}$ budget while preserving the decomposition, we first clip the aggregated perturbation and then apply a shared per-pixel rescaling to both components:
\begin{equation}
\label{eq:linf_rebudget}
    \bar{\delta}_t = \mathrm{clip}(\delta_t, -\epsilon, \epsilon), \quad
    s_t = \frac{\bar{\delta}_t}{\delta_t + \eta},
\end{equation}
\begin{equation}
\label{eq:component_rebudget}
    \delta_t^{L} \leftarrow \delta_t^{L} \odot s_t, \quad
    \hat{\delta}_t^{H} \leftarrow \hat{\delta}_t^{H} \odot s_t,
\end{equation}
\begin{equation}
\label{eq:pixel_clip}
    x_{t+1} = \mathrm{clip}(x + \delta_t^{L} + \hat{\delta}_t^{H}, 0, 1),
\end{equation}
where $\eta=10^{-12}$ avoids division by zero and $\odot$ denotes element-wise multiplication.
In practice, this shared rescaling approximately preserves the intended orthogonality while satisfying the $\epsilon$-ball constraint.

\paragraph{Plug-and-Play Integration}
Season is implemented as a stateful wrapper around an arbitrary base transfer attack.  
At each iteration, a gradient estimator $\Phi_{\text{base}}$ (e.g., MI-FGSM, DIM, TIM, VT, Admix) is queried to produce the surrogate gradient $g_t$ possibly after applying input transformations or neighborhood sampling.  
For look-ahead methods such as NI-FGSM and PI-FGSM, Season follows their original definitions by computing gradients at a shifted point determined by the aggregated momentum, while still using the dual-branch refinement described above.  
In all cases, Season \emph{only} replaces the original single-stream momentum accumulation and update rule with its dual-frequency, mask-guided, and orthogonally refined counterpart, leaving all other hyper-parameters and transformation schedules of the base attack unchanged.  
This design introduces only a small constant-factor overhead (a Gaussian filtering and a pre-computed saliency map per image) and allows Season to serve as a generic plug-and-play module for enhancing cross-architecture transferability.

\section{EXPERIMENTS}
\label{sec:experiments}

\subsection{Experimental Setup}
\noindent\textbf{Dataset and Models.}
We follow the standard untargeted transfer-based black-box setting~\cite{papernot2017practical} on ImageNet-1k.
We randomly sample 5,000 validation images that are correctly classified by the source (surrogate) model (\textbf{ResNet-50}).
Transferability is evaluated on eight unseen targets: ResNet-101 (RN-101), DenseNet-121 (DN-121), VGG-19, MobileNet-V2 (MB-V2),
GoogLeNet (Google), ViT-B/16 (ViT-B)~\cite{dosovitskiy2020image}, Swin-T, and MLP-Mixer-B/16 (Mixer-B).
This selection covers diverse architectures including CNNs, Transformers, and MLPs.

\noindent\textbf{Attacks and Configuration.}
We evaluate Season on top of eight representative transfer attacks:
MI-FGSM~\cite{dong2018boosting}, NI-FGSM~\cite{lin2019nesterov}, PI-FGSM~\cite{gao2020patch}, DI-FGSM~\cite{xie2019improving},
TI-FGSM~\cite{dong2019evading}, SI-FGSM~\cite{lin2019nesterov},
VT (Variance Tuning; VMI-FGSM)~\cite{wang2021enhancing}, and Admix~\cite{wang2021admix}.
To ensure fair comparison, all baselines follow the unified TPAMI benchmark settings~\cite{zhao2025revisiting}.
Unless otherwise stated, attacks are run for $T=10$ iterations with $\epsilon=16/255$ and step size $\alpha=2/255$.
For Season, we set the saliency percentile $\rho=0.15$, suppression factor $\lambda=0.7$, and $\sigma=1.0$ (kernel size $k=5$) by default.

\noindent\textbf{Metric.}
We report the Transfer Success Rate (TSR), defined as the percentage of adversarial examples that successfully fool the target model.

\subsection{Main Results: Comparison with Strong Baselines}
\label{subsec:main-results}
We first evaluate the universal effectiveness of Season. Table~\ref{tab:main_results} reports the TSR across all source-target pairs.


\paragraph{Universal Boosting}
Season improves transferability across all eight baselines and in almost all individual target cases.
On average, Season boosts the mean TSR by \textbf{6.6\%}.
Notably, for TI-FGSM, Season achieves a substantial gain of \textbf{+15.5\%} (53.5\% $\to$ 69.0\%). Since TI smooths gradients to achieve translation invariance, it inadvertently suppresses high-frequency signals. Season successfully reclaims this textural attack capability via our dual-frequency decomposition, suggesting that \emph{spatial smoothing and spectral refinement are complementary}.

\paragraph{Bridging the Architecture Gap}
Transferring from CNNs to ViTs remains a key challenge. Season demonstrates remarkable gains in this setting. For instance, on the difficult ViT-B/16 target, Season improves SI-FGSM from 30.6\% to \textbf{40.5\%} (+9.9\%) and Admix from 38.7\% to \textbf{42.2\%}.
This supports our core motivation that by decoupling the optimization, Season allows the attack to exploit CNN textures (via the High-Freq branch) without corrupting the structural cues (via the Low-Freq branch) essential for ViTs.

\paragraph{Best Overall Performance under the Unified TPAMI Setting}
Comparing against strong baselines under the unified TPAMI setting, Season-DI achieves the highest average TSR of \textbf{83.8\%}, while Season-VT reaches \textbf{82.8\%}, significantly outperforming their vanilla counterparts. This demonstrates Season's plug-and-play capability to enhance even the strongest existing methods without requiring complex hyper-parameter tuning.

\paragraph{Robustness to Surrogate Choice and Attack Budgets}
To address concerns about using a single surrogate and a single budget configuration, we further evaluate Season under five CNN surrogates, two representative baselines (MI-FGSM/DI-FGSM), and two $\ell_\infty$ budgets.
Table~\ref{tab:robustness} reports the average transfer success rate (Avg TSR) over the eight target models, where we use $T=10$ and set the step size as $\alpha=\epsilon/T$ for each budget.

\begin{table}[t]
\centering
\caption{Robustness study across diverse CNN surrogates and $\ell_\infty$ budgets.
We report Avg TSR (\%) over the eight target models on the fixed 1,024-image subset with $T=10$ and set the step size as $\alpha=\epsilon/T$.
$\Delta$ denotes the change in Avg TSR brought by Season (in percentage points, pp).}
\label{tab:robustness}
\setlength{\tabcolsep}{4pt}
\scriptsize
\begin{tabular}{lcccccc}
\toprule
Surrogate & $\epsilon$ & $T$ & Attack & Baseline & +Season & $\Delta$ (pp) \\
\midrule
DN-121  & 8/255  & 10 & DI-FGSM & 64.1 & 64.3 & \textbf{+0.2} \\
DN-121  & 8/255  & 10 & MI-FGSM & 53.2 & 53.5 & \textbf{+0.3} \\
DN-121  & 16/255 & 10 & DI-FGSM & 78.4 & 78.1 & -0.3 \\
DN-121  & 16/255 & 10 & MI-FGSM & 67.7 & 68.8 & \textbf{+1.1} \\
\midrule
GoogLeNet  & 8/255  & 10 & DI-FGSM & 54.2 & 54.6 & \textbf{+0.4} \\
GoogLeNet  & 8/255  & 10 & MI-FGSM & 40.8 & 41.5 & \textbf{+0.8} \\
GoogLeNet  & 16/255 & 10 & DI-FGSM & 71.1 & 70.5 & -0.6 \\
GoogLeNet  & 16/255 & 10 & MI-FGSM & 55.6 & 57.2 & \textbf{+1.6} \\
\midrule
MB-V2   & 8/255  & 10 & DI-FGSM & 61.5 & 62.5 & \textbf{+1.0} \\
MB-V2   & 8/255  & 10 & MI-FGSM & 46.2 & 47.8 & \textbf{+1.7} \\
MB-V2   & 16/255 & 10 & DI-FGSM & 77.4 & 77.6 & \textbf{+0.2} \\
MB-V2   & 16/255 & 10 & MI-FGSM & 60.5 & 62.8 & \textbf{+2.3} \\
\midrule
RN-50   & 8/255  & 10 & DI-FGSM & 66.0 & 67.2 & \textbf{+1.3} \\
RN-50   & 8/255  & 10 & MI-FGSM & 51.0 & 51.9 & \textbf{+0.9} \\
RN-50   & 16/255 & 10 & DI-FGSM & 79.6 & 80.4 & \textbf{+0.9} \\
RN-50   & 16/255 & 10 & MI-FGSM & 65.8 & 68.0 & \textbf{+2.2} \\
\midrule
VGG-19  & 8/255  & 10 & DI-FGSM & 55.0 & 55.7 & \textbf{+0.8} \\
VGG-19  & 8/255  & 10 & MI-FGSM & 41.9 & 44.9 & \textbf{+3.0} \\
VGG-19  & 16/255 & 10 & DI-FGSM & 68.4 & 69.2 & \textbf{+0.8} \\
VGG-19  & 16/255 & 10 & MI-FGSM & 55.7 & 59.3 & \textbf{+3.6} \\
\bottomrule
\end{tabular}
\end{table}

\subsection{Ablation Study: Dissecting the Mechanism}
\label{subsec:ablation}
To validate our design choices, we conduct a component-wise ablation using MI-FGSM on a subset of 1,024 images. Table~\ref{tab:ablation} summarizes the results on representative targets.


\paragraph{Effect of Dual-Frequency Decomposition}
Comparing (C) vs.\ (A), simply decomposing the gradient boosts ViT transferability by \textbf{+4.0\%} (29.9\% $\to$ 33.9\%), while CNN performance remains stable. This suggests that \emph{explicitly isolating the low-frequency component} is beneficial for preserving shape cues, as standard coupled gradients tend to be texture-dominated.

\paragraph{Effect of Low-Saliency Guidance}
Comparing (D) vs.\ (C), applying the low-saliency mask to the high-frequency branch significantly improves CNN transferability (+2.2\% on RN-101). This supports our hypothesis that CNNs are highly vulnerable to background texture perturbations.
As qualitatively visualized in Fig.~\ref{fig:teaser}(b), Season tends to push noise into background regions, thereby maintaining object structure. This strategy yields a better overall cross-architecture trade-off.
However, enabling the mask can introduce a small trade-off on ViTs (E: 36.13\% $\rightarrow$ F: 35.05\%) while improving CNN/MLP targets and achieving the best average TSR.

\paragraph{Effect of Orthogonal Refinement}
Orthogonal projection is crucial for preserving structural cues and improving ViT transfer. Compared to Dual-only (C), adding orthogonal refinement (E) achieves the highest TSR on ViT-B/16 (33.85\% $\rightarrow$ 36.13\%).
When the mask is enabled, orthogonal refinement (F vs. D) further improves RN-101 (91.83\% $\rightarrow$ 93.23\%) and maintains comparable ViT performance (34.81\% $\rightarrow$ 35.05\%), at a minor cost on Mixer-B (42.42\% $\rightarrow$ 42.01\%), leading to the best overall average TSR (56.35\% $\rightarrow$ 56.76\%).
This empirical evidence supports our design in Sec.~III-C, where orthogonal refinement is introduced to mitigate interference between textural and structural updates.

\begin{table}[t]
\centering
\caption{Ablation study on Season components. Transfer success rate (\%) on representative targets.}
\label{tab:ablation}
\resizebox{\columnwidth}{!}{%
\begin{tabular}{l|ccc|ccc|c}
\hline
\textbf{Method} & \textbf{Dual} & \textbf{Mask} & \textbf{Ortho} & \textbf{RN-101} & \textbf{ViT-B/16} & \textbf{Mixer-B} & \textbf{Avg.} \\
\hline
A. Baseline (MI) & - & - & - & 89.53 & 29.89 & 39.6 & 53.01 \\
B. Single + Mask & - & \checkmark & - & 89.66 & 30.97 & 40.88 & 53.84 \\
C. Dual Only & \checkmark & - & - & 89.66 & 33.85 & 40.88 & 54.8 \\
D. Dual + Mask & \checkmark & \checkmark & - & 91.83 & 34.81 & \textbf{42.42} & 56.35 \\
E. Dual + Orth & \checkmark & - & \checkmark & 91.44 & \textbf{36.13} & 41.61 & 56.39 \\
\textbf{F. Full Season} & \checkmark & \checkmark & \checkmark & \textbf{93.23} & 35.05 & 42.01 & \textbf{56.76} \\
\hline
\end{tabular}%
}
\end{table}

\section{Conclusion}

In this paper, we addressed the critical challenge of cross-architecture adversarial transferability, identifying the \emph{curse of coupled gradients} as a primary cause for the poor transfer from CNN surrogates to Vision Transformers. To overcome this, we proposed \textbf{Season}, a universal spectrum-aware orthogonal gradient refinement framework. By explicitly decoupling the optimization process, Season coordinates low-frequency structural updates and high-frequency textural perturbations through a geometric orthogonality constraint, effectively resolving the conflict between shape-biased and texture-biased features. Furthermore, we introduced a low-saliency guidance strategy that exploits CNN vulnerabilities while preserving visual stealthiness.

Extensive evaluations on ImageNet demonstrate that Season acts as a powerful plug-and-play module, consistently boosting the transferability of eight representative attacks against diverse CNN, ViT, and MLP targets. Notably, Season successfully reclaims the high-frequency attack capability suppressed in smoothing-based methods (e.g., TI-FGSM) and achieves the best average TSR among the evaluated Season-enhanced variants when combined with Variance Tuning. Our work reveals the importance of spectral and geometric disentanglement in adversarial optimization, providing a robust baseline for evaluating the security of heterogeneous deep learning systems.

\section*{Acknowledgment}
The authors thank Wuhan LightRead Intelligent Technology Co., Ltd. for providing partial technical-environment and financial support for this work. This support facilitated experiment deployment, computing-resource preparation, and empirical evaluation throughout the project.

\bibliographystyle{IEEEbib}
\bibliography{icme2026references}

\end{document}